\documentclass[sigconf]{acmart}
\AtBeginDocument{%
  \providecommand\BibTeX{{%
    \normalfont B\kern-0.5em{\scshape i\kern-0.25em b}\kern-0.8em\TeX}}}

\setcopyright{acmcopyright}
\copyrightyear{2018}
\acmYear{2018}
\acmDOI{10.1145/1122445.1122456}

\acmConference[AdvML '21]{AdvML '21: Workshop on Adversarial Learning Methods for Machine Learning and Data Mining}{August 15, 2021}{Singapore (virtual)}
\acmBooktitle{AdvML '21: Workshop on Adversarial Learning Methods for Machine Learning and Data Mining, August 15, 2021, Singapore (virtual)}

\usepackage{microtype}
\usepackage{graphicx}
\usepackage{amsmath}
\usepackage{booktabs} 
\usepackage{appendix}
\usepackage{enumitem, kantlipsum}
\usepackage{multirow}
\usepackage{nicefrac}
\usepackage{subcaption} 
\usepackage{wrapfig}

\begin{document}

\title{STRATA: Simple, Gradient-Free Attacks for Models of Code}

\author{Jacob M. Springer}
\authornote{Both authors contributed equally to this research.}
\email{jacmspringer@gmail.com}
\affiliation{%
  \institution{MIT}
  \city{Cambridge}
  \state{MA}
  \country{USA}
}

\author{Bryn Marie Reinstadler}
\authornotemark[1]
\email{brynr@mit.edu}
\affiliation{%
  \institution{MIT}
  \city{Cambridge}
  \state{MA}
  \country{USA}
}

\author{Una-May O'Reilly}
\email{unamay@csail.mit.edu}
\affiliation{%
  \institution{MIT}
  \city{Cambridge}
  \state{MA}
  \country{USA}
}

\renewcommand{\shortauthors}{Springer et al.}

\newcommand{\sname}[1]{{\small \textsc{#1}}}
\newcommand{\codetoseq}{\sname{code2seq}}
\newcommand{\bilstm}{\sname{BiLSTM}}
\newcommand{\javalarge}{\sname{java-large}}
\newcommand{\javamed}{\sname{java-med}}
\newcommand{\javasmall}{\sname{java-small}}
\newcommand{\pydata}{\sname{python150k}}

\begin{abstract}
    Neural networks are well-known to be vulnerable to imperceptible perturbations in the input, called adversarial examples, that result in misclassification. Generating adversarial examples for source code poses an additional challenge compared to the domains of images and natural language, because source code perturbations must retain the functional meaning of the code. We identify a striking relationship between token frequency statistics and learned token embeddings: the L2 norm of learned token embeddings increases with the frequency of the token except for the highest-frequnecy tokens. We leverage this relationship to construct a simple and efficient gradient-free method for generating state-of-the-art adversarial examples on models of code. Our method empirically outperforms competing gradient-based methods with less information and less computational effort.
\end{abstract}




\maketitle

\section{Introduction}

Although machine learning has been shown to be effective at a wide variety of tasks across computing, statistical models are susceptible to \textit{adversarial examples} \citep{szegedy_intriguing_2014}. Researchers have developed effective techniques for adversarial example generation in the image domain \citep{goodfellow_explaining_2015, moosavi-dezfooli_universal_2017, papernot2016limitations} and in the natural language domain \citep{alzantot_generating_2018, ebrahimi_hotflip_2018, michel_evaluation_2019, cheng_seq2sick_2020, belinkov_synthetic_2018, papernot_crafting_2016}, although work in the source code domain is less extensive (see Related Work). We focus on generating adversarial examples for neural program analyzers \citep{allamanis2016convolutional, iyer2016summarizing, allamanis2015bimodal, alon_code2seq_2019, alon2019code2vec}, which draw highly successful techniques from the domain of natural language processing. Typically, neural models of code learn distributed representations of observed tokens, called token embeddings \citep{mikolov2013distributed, mikolov2013efficient, allamanis2016convolutional, iyer2016summarizing, allamanis2015bimodal, alon_code2seq_2019, alon2019code2vec}.

We ask the question: \textit{are there interesting statistical relationships between training dataset distributions and learned token embeddings that can be leveraged to construct effective adversarial examples?} We find that source code embeddings consistently exhibit a striking pattern: tokens with a low frequency in the training data correspond with low-L2-norm embeddings, and tokens with a high frequency in the training data correspond with high-L2-norm embeddings. Past an inflection point, very-high-frequency tokens correspond with low-L2-norm embeddings. We use this relationship to propose the Simple Trained Token Attack (\sname{strata}), a state-of-the-art \textit{variable replacement adversarial attack} technique in \codetoseq{} \citep{alon_code2seq_2019, rabin_evaluation_2020, ramakrishnan_semantic_2020}. In particular, we find that there is a tight correlation between the L2 norm of a token embedding and the impact the token can have on the model when used in a variable replacement adversarial attack. We report that applying a simple token replacement attack using tokens identified by a high L2 norm can outperform other state-of-the-art attacks on \codetoseq{}. In addition, we can leverage the relationship between the token embedding L2 norm and the token frequencies to generate similarly effective black-box attacks. We proceed as follows: first, we empirically demonstrate the existence of a correlation between token frequency in the training data and learned token embedding L2 norm, then, based on this relationship, we construct effective, computationally efficient adversarial examples for the \codetoseq{} model.


\section{High frequency tokens from training set exhibit higher L2 norm}

\begin{figure*}[ht!]
    \centering
    \begin{subfigure}[t]{.24\textwidth}
    \includegraphics[width=\textwidth]{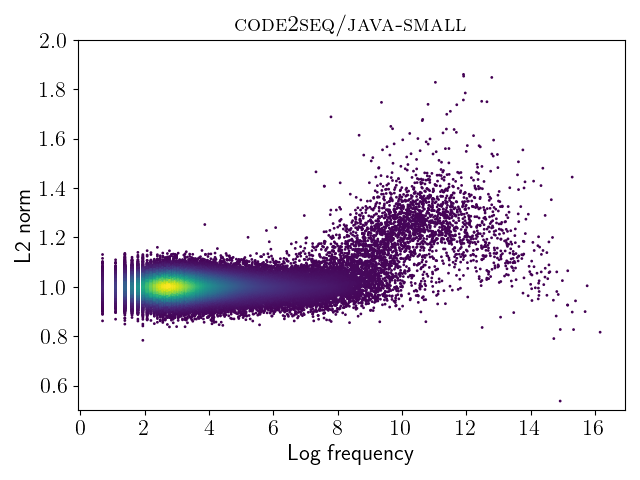}
    \end{subfigure}
    \begin{subfigure}[t]{.24\textwidth}
    \includegraphics[width=\textwidth]{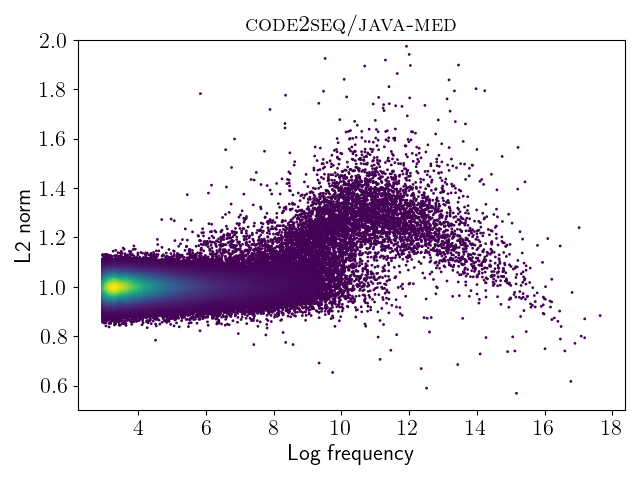} 
    \end{subfigure}
    \begin{subfigure}[t]{.24\textwidth}
    \includegraphics[width=\textwidth]{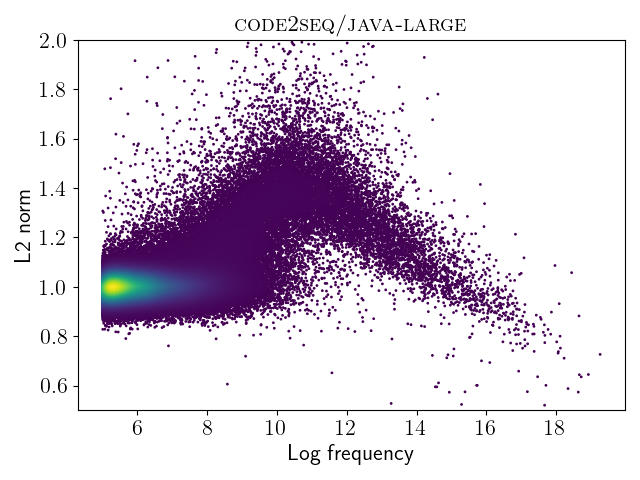} 
    \end{subfigure}
    \begin{subfigure}[t]{.24\textwidth}
    \includegraphics[width=\textwidth]{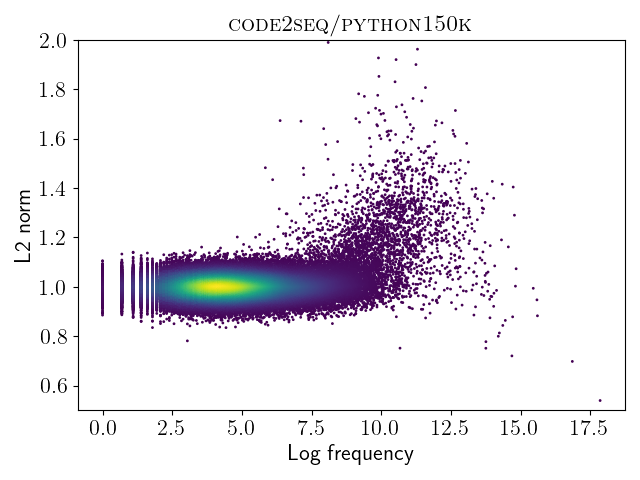} 
    \end{subfigure}
    \caption{L2 norm of trained token embeddings by frequency of token in training dataset. Color represents the approximate density of tokens for a given frequency and embedding L2 norm. A lighter color corresponds with a higher density.}
    \label{fig:l2_frequency}
\end{figure*}

To demonstrate a statistical relationship that gives insight into the interpretation of learned distributed representations of tokens, we assay the relationship between the frequency of a token within the training set and the L2 norm of the post-training embedding vector in the \codetoseq{} model trained on different datasets.

\paragraph{\codetoseq{}.} This model, developed by \citet{alon_code2seq_2019}, is an encoder-decoder model inspired by \sname{seq2seq} \citep{sutskever2014sequence}; it operates on code rather than natural language. \sname{code2seq} is the state-of-the-art model for code summarization, and therefore it represents a good target for adversarial attacks and adversarial training. The model is tasked to predict method names from the source code body of a method. The model considers both the structure of an input program's Abstract Syntax Trees (ASTs) as well as the tokens corresponding to identifiers such as variable names, types, and invoked method names. To reduce the vocabulary size, identifiers are split into tokens by commonly-used delimiters such as camelCase and under\_scores. \sname{code2seq} then encodes these tokens into distributed embeddings. In this paper, we distinguish between \textit{token} embeddings and \textit{identifier} embeddings. \textit{Token} embedding vectors are trained model parameters. \textit{Identifier} embedding vectors are computed as a sum of the embeddings of the constituent tokens. If the identifier contains more than five tokens, only the first five are summed, as per the \sname{code2seq} architecture. The full description and architecture of the \sname{code2seq} model is given in the original paper by \citet{alon_code2seq_2019}.

\paragraph{Datasets.}
In this work, we use four datasets:
\begin{enumerate}[noitemsep,nolistsep]
    \item Java-small (\javasmall{}, 700k examples),
    \item Java-medium (\javamed{}, 4M examples),
    \item Java-large (\javalarge{}, 16M examples),
    \item Python150k (\pydata{}, 150k examples)
\end{enumerate}

The Java datasets are proposed by \citet{alon_code2seq_2019} and the Python dataset is proposed by \citet{raychev2016probabilistic}. None of the datasets overlap. We disambiguate the trained \sname{code2seq} models for each datasets by denoting them \sname{code2seq/java-small}, \sname{code2seq/java-med}, \sname{code2seq/java-large}, and \sname{code2seq/python150k} for models trained on \javasmall{}, \javamed{}, \javalarge{}, and \pydata{} respectively. 

For each of the models we examine (\codetoseq{} trained on \javasmall{}, \javamed{}, \javalarge{}, \pydata{}), we plot the relationship between token frequency and the L2 norm of the associated embedding, and find a striking relationship: low-frequency tokens correspond with low L2 norm embeddings, high-frequency tokens correspond with high L2 norm embeddings, and, past an inflection point, very-high-frequency tokens have low L2 norm embeddings (Figure~\ref{fig:l2_frequency}). The tokens with highest L2 norm token embeddings occupy a ``Goldilocks zone'' such that the frequency of the token is ideal: not too low and not too high. This is consistent with our intuitive notion of the information carried by each token. Low-frequency tokens do not occur enough to carry significant information. The highest frequency tokens appear so frequently that their presence does not indicate meaningful information. Thus, tokens that carry the most information should be frequent enough as to have enough information to draw conclusions, but not so frequently that it cannot be used to distinguish between different contexts.

\section{Token impact on model accuracy under variable replacement attack can be measured by L2 norm}

We next establish an interesting implication of the identified relationship between embedding L2 norm and token frequency: in the \codetoseq{} model, the impact of a token on the accuracy of a model under a \textit{variable replacement attack} correlates with the L2 norm of the token. A \textit{variable replacement attack} is a replacement of a single variable name with a different valid but adversarial name, with the goal of changing the prediction of the model using this small change. We ask: \textit{can we can construct highly effective variable-replacement based adversarial examples by exchanging variable names for tokens with high L2 norm embeddings?} Furthermore, \textit{will the observed frequency-L2 relationship allow us to formulate black-box variable replacement attacks without requiring information about the L2 norm of the embedding vectors?} We find that, for both questions, the answer is yes.

\subsection{The variable replacement attack}

\begin{table*}[ht!]
\centering
\caption{Evaluation of F1 scores of the \sname{strata} on our \codetoseq{} models, using vocabularies and replacement strategies as described in the Methods section. The baseline refers to the performance of the model on the original test dataset. For the Java models, samples were drawn from the \javasmall{} testing set such that each method included a local variable to perturb. All top-$n$ scores use optimal values of $n$ proposed in Table~\ref{tab:topn}.}
\label{tab:strata_attack_stats}
\setlength\tabcolsep{3pt}
\begin{tabular}{c|c|ccc|ccc|ccc} \toprule
& Baseline & \multicolumn{3}{c|}{All} & \multicolumn{3}{c|}{Top-$n$ by L2 norm} & \multicolumn{3}{c}{Top-$n$ by Frequency} \\ \midrule
 &      & \textit{single}      & \textit{5-diff}    & \textit{5-same}    & \textit{single}      & \textit{5-diff}    & \textit{5-same}   & \textit{single}        & \textit{5-diff}      & \textit{5-same}     \\ \midrule
\sname{code2seq/java-small}  & .369 & .381   & .350   & .310   & .362   & .263   & \textbf{.214}  & .372     & .284     & \textbf{.231}    \\
\sname{code2seq/java-med}  & .564 & .548   & .531   & .492   &  .513  & .416   & \textbf{.375} &   .513   &    .385  & \textbf{.345}  \\
\sname{code2seq/java-large}  & .608 & .536   & .547   & .488   & .542   & .396   & \textbf{.360}  & .548     & .427     & \textbf{.388}   \\
\sname{code2seq/python150k}  & .313 & .274   & .250   & .209   & .249   & .198   & \textbf{.153}  & .256     & .211     & \textbf{.172}   \\
\bottomrule
\end{tabular}
\end{table*}

\begin{table}[ht!]
\centering
\caption{Effectiveness of targeted attacks on \sname{code2seq}. All attacks employ the \textit{5-same} replacement strategy.}
\label{tab:targeted_attack}
\begin{tabular}{lll} \toprule 
Model                & Perturbation               & \% success \\ \midrule
\sname{code2seq/java-large} & L2, top 6k          & \textbf{37.1}                             \\
\sname{code2seq/java-large}& Freq, top 10k & 35.6                                 \\
\sname{code2seq/java-large}& All          & 3.9                                 \\ \midrule
\sname{code2seq/java-med} & L2, top 3k    &              \textbf{43.8}               \\
\sname{code2seq/java-med} & Freq, top 3k   &          39.1                        \\
\sname{code2seq/java-med} & All &                        1.4                    \\ \midrule
\sname{code2seq/java-small} & L2, top 1k    & \textbf{58.7}                             \\
\sname{code2seq/java-small} & Freq, top 1.8k  & 52.8                                 \\
\sname{code2seq/java-small} & All &            2.1                                \\\bottomrule
\end{tabular}
\end{table}

Traditional adversarial attacks on discrete spaces involve searching the discrete space for semantically similar perturbations that yield a misclassification. Searching the space of all possible valid discrete changes in source code is often intractable or even impossible \citep{rice1953classes}. However, restricting the search space to a set of predefined operations that maintain semantic invariance can reduce the search space. Gradient-based attacks on the embedding space may also be used in order to optimize the search of the space itself \citep{yefet_adversarial_2020, ramakrishnan_semantic_2020}. However, gradient-based attacks are computationally expensive and rely heavily on knowledge of the exact parameters of the model.

We eliminate the need for an expensive search operation by performing a local variable name replacement attack using high-impact tokens. The observed frequency-L2 relationship motivates two ways to identify high-impact tokens: first, we should consider tokens with a high L2 norm embeddding, and second, since we have established a correlation between high L2 norm embeddings and token frequency, we can use tokens selected on the basis of high frequency in the training set instead.\footnote{It is noted that very-high-frequency tokens have lower L2 norms. However, since very-high frequency tokens with low L2 norm are relatively rare, they are not expected to affect results significantly.} 

Thus, in our experiments, we choose high-impact tokens in three different ways:
\begin{enumerate}[noitemsep,nolistsep]
    \item \textit{All:} Contains all tokens. Note that the number of tokens varies by dataset (Table~\ref{tab:topn});
    \item \textit{Top $n$ by L2 norm} contains tokens for which their L2 norm embedding vectors are the $n$ highest;
    \item \textit{Top $n$ by frequency} contains only the $n$ tokens which occur in the training data with highest frequency.
\end{enumerate}

We propose three strategies to generate an identifier with a large L2-norm embedding:

\begin{enumerate}[noitemsep,nolistsep]
    \item \textit{single:} pick a single token to use as the identifier;
    \item \textit{5-diff:} pick five different (not necessarily unique) tokens and concatenate them to form a compound identifier, which will have a higher expected L2 norm than \textit{single};
    \item \textit{5-same:} pick a single token, and concatenate the token five times with itself to form a compound identifier, which will have the largest expected L2 norm, by the triangle inequality\footnote{The triangle inequality states that $\lVert \mathbf{x} + \mathbf{y} \rVert \leq \lVert \mathbf{x} \rVert + \lVert \mathbf{y} \rVert$, and is equal (and thus maximized) when $\mathbf{x}$ and $\mathbf{y}$ are colinear, which occurs when $\mathbf{x} = \mathbf{y}$. This is easily generalized to five vectors.}.
\end{enumerate}

First, we wish to construct untargeted adversarial attacks, i.e., attacks that aim to simply change the model prediction. For a given method in the source code, \sname{strata} will generate an untargeted perturbation as follows:
\begin{enumerate}[noitemsep,nolistsep]
    \item Select one random local variable $v$;
    \item Choose an adversarial token $v^*$ appropriately, using the chosen concatenation strategy (\textit{single}, \textit{5-diff}, or \textit{5-same}). For white-box attacks, choose each token from a high-L2-norm vocabulary (top-$n$ by L2 norm). For black-box attacks, choose each token with sufficiently high frequency (top-$n$ by frequency). We discuss the optimal cutoff values ($n$) for L2 and frequency in the appendix.
    \item Replace $v$ with $v^*$.
\end{enumerate}

For attacks on the Python dataset, since determining whether a variable is local or non-local is not always possible by looking at only the body of the method, the attack treats all variables as local.

To perform targeted attacks, in which the output includes a particular token $t$, we perform the same steps as for the untargeted attack, with the exception of choosing $v^*$ to be a \textit{5-same} concatenation of $t$.

\section{Results}

To confirm that our attack strategy is successful, we compare the F1 score of each Java model on the \javasmall{} testing dataset baseline and on the dataset perturbed by each type of adversarial perturbation (Table~\ref{tab:strata_attack_stats}). For the Python model, we compute F1 scores on the \pydata{} testing dataset and the associated perturbations of the dataset. Lower F1 scores correspond with better attacks. The attack performance is optimal when using the \textit{5-same} replacement strategy and when tokens are selected by the optimal top-$n$ by L2 norm vocabulary, but the attack also performs well when tokens are selected from the optimal top-$n$ by frequency vocabulary. Thus, we confirm our hypothesis that we can improve the effectiveness of adversarial examples by selecting replacement identifiers with high token embedding L2 norm. Similarly, we can use the token frequency in the training dataset as an effective proxy to identify high-impact tokens. In fact, the F1 scores of the white-box L2-norm attack and the black-box frequency attack are similar, suggesting that the black-box method in which we choose adversarial tokens based on frequency alone can approximate the white-box attack.

Surprisingly, the F1 score of \sname{code2seq/java-small} increased for random and frequency-based adversarial perturbations constructed with the \textit{single} concatenation strategy, suggesting that \sname{code2seq/java-small} relies less on variable names for classification than \sname{code2seq/java-med} or \sname{code2seq/java-large}. The attacks on \sname{code2seq/python150k} were also highly effective, although the baseline accuracy for that model was already lower than the rest of the models.

\subsection{Targeted attacks}

We also demonstrate that when tokens with a large L2 norm are used for variable replacement, we can construct targeted adversarial attacks. The correspondence between embedding L2 norm and token frequency allows us to construct an analogous black-box attack in which we select high-frequency tokens rather than high embedding L2 norm tokens for variable replacement.

To assay the effectiveness of targeted attacks, we perform targeted attacks that target three different token vocabularies: (1) all valid tokens, (2) the optimized L2 vocabulary, and (3) the optimized frequency vocabulary, where vocabularies are optimized for \sname{code2seq/java-small}, \sname{code2seq/java-med}, or \sname{code2seq/java-large} appropriately. We define a particular attack as successful if the selected token is included in the output. We measure the percent of successful attacks, thus computing an aggregate effectiveness of targeted attacks (Table~\ref{tab:targeted_attack}).

Table~\ref{tab:targeted_attack} reveals that \sname{code2seq} is especially vulnerable to targeted attacks performed on high-impact tokens. The black-box (frequency) attack performs similarly to the white-box (L2) attack.

\section{STRATA outperforms similar attacks}

We compare our work with the gradient-based adversarial perturbations proposed by \citet{ramakrishnan_semantic_2020}, in which they attack a \sname{code2seq/java-small} model. We consider the variable replacement, print statement insertion, try-catch statement insertion, and worst-case single transformation attacks for our comparison and report the performances from their paper. For a fair comparison, we include the examples in the training set that do not include local variables, and thus we do not even attempt to perturb them, hence why the F1 scores of the adversarial examples are larger than as reported in Table~\ref{tab:strata_attack_stats}. We find that \sname{strata} outperforms all attacks performed by \citet{ramakrishnan_semantic_2020} except for the worst-case transformation, which is inherently a larger transformation than our variable-replacement attack. Since our technique is gradient-free, it is substantially more computationally efficient and does not rely on the parameters of the entire model, even in the white-box case.

\begin{table}[t]
\centering
\caption{Comparison of F1 scores our method to gradient-based attacks described by \citet{ramakrishnan_semantic_2020}. Both our model and the \citet{ramakrishnan_semantic_2020} model are trained on \javasmall{}. We perform \textit{5-same} attacks.}
\label{tab:ramakrishnan_comparison}
\begin{tabular}{lcc}
\toprule 
Adversarial attack  & F1  & \% of baseline  \\ \midrule
\multicolumn{3}{c}{\citet{ramakrishnan_semantic_2020} attacks} \\
Baseline  & .414   & 100       \\
Variable Replacement & .389  & 93.9  \\
Try-catch Statement Insertion & .336 & 81.2 \\
Print Statement Insertion & .312   & 75.3  \\
Worst-case Transformation & .246 & 59.4 \\
\midrule
\multicolumn{3}{c}{\sname{strata}} \\
Baseline & .425   & 100       \\
Top 1k by L2  & \textbf{.316} &  \textbf{74.4} \\
Top 1.8k by frequency  & .328 & 77.2 \\
\bottomrule 
\end{tabular}
\end{table}

\section{Related work}

\paragraph{Token embeddings in source code and natural language}
Distributed representations of words and tokens have been proposed as a method to effectively train deep learning models on natural language and have been shown to have some compositional semantic structure \citep{collobert2008unified, bengio2003neural, mikolov2013distributed, mikolov2013efficient}. Similarly, source code token embeddings have been used for various tasks, including program repair, program analysis, predicting code correctness, and deep learning for code summarization \citep{harer2018automated, henkel2018code, white2019sorting, azcona2019user2code2vec, chen2018remarkable, alon_code2seq_2019, alon2019code2vec}.

\paragraph{Adversarial examples for models of code} There has been work on adversarial examples and model robustness in other domains \citep{szegedy_intriguing_2014, goodfellow2014explaining, madry2018towards, shafahi2019adversarial, wong2019fast}. \citet{allamanis2018survey} provide a comprehensive survey of prior research on models of code. Several papers develop techniques for generating adversarial examples on models of source code: \citet{quiring_misleading_2019} perform adversarial attacks on source code by applying a Monte-Carlo tree search over possible transformations, and \citet{zhang_generating_2020} apply Metropolis-Hastings sampling to perform identifier replacement to create adversarial examples. \citet{bielik2020adversarial} improve on the adversarial robustness of models of code by developing a model that can learn to abstain if uncertain. 

\paragraph{Adversarial examples for \sname{code2vec} and \sname{code2seq}} \citet{yefet_adversarial_2020} generate both gradient-based targeted and untargeted adversarial examples for \sname{code2vec} \citet{alon2019code2vec}. Most directly related to our paper, \citet{ramakrishnan_semantic_2020} perform gradient-based adversarial attacks and adversarial training on \sname{code2seq}. \citet{rabin_evaluation_2020} evaluate the robustness of \sname{code2seq} to semantic-preserving transformations.

\paragraph{Targeted attacks on models of code}
As previously noted, \citet{yefet_adversarial_2020} propose a method for targeted adversarial examples for \sname{code2vec} \citep{alon2019code2vec}. To our knowledge, at the time of writing, no other paper performs targeted attacks on \sname{code2seq}.

\section{Conclusion}

In this work, we presented \sname{STRATA}, a simple, gradient-free method for generating adversarial examples in discrete space that can be used to help build robustness in models of code. Because the L2 norm of an embedding vector can be approximated by the frequency in the training data, \sname{STRATA} can be used to generate gradient-free white- and black-box attacks in a targeted and untargeted setting. We base our adversarial attack generation method on the observed relationship between the L2 norm of token embeddings and the frequency with which the tokens appear in the dataset. In particular, we find that low-frequency and very-high-frequency tokens exhibit low-L2-norm embeddings while high-frequency tokens exhibit high-L2-norm embeddings.

We believe that the relationship between token frequency and embedding L2 norm is general, and should extend to other neural models of code. In fact, we expect our result to extend to neural models of natural language as well, due to the similarities between the neural architectures and training techniques of natural languages models and models of code.

We speculate that our observation that tokens with high embedding L2 norm can be effective candidates for variable replacement may help drive future work aiming to improve model robustness and defend against adversarial attacks. For example, a defender might use these statistics to help determine whether or not a particular variable name is adversarial. In addition, we expect the relationship between the frequency of tokens and the L2 norm of their embeddings to have broad implications that will drive insights into the underlying nature of embeddings. We expect that there many applications of our result which we do not explore in this paper.


\bibliographystyle{ACM-Reference-Format}
\bibliography{main}

\appendix

\onecolumn

\section{Implementation}
\label{subsection:implementation}

We assayed the effectiveness of the proposed token selection strategy by applying a simple local variable token replacement attack. We select a single locally-declared variable at random and rename it with an adversarial identifier which does not conflict with another name. Because the initial variable is locally declared, we know that changing the name will have no effect elsewhere in the program, and will have no behavioral effect. The token replacement, however, can effectively attack \codetoseq{}; example shown in the Appendix.

We use the \javasmall{} test set as testing data for \sname{code2seq/java-small}, \sname{code2seq/java-med}, and \sname{code2seq/java-large}. This test set consists of approximately 57,000 examples; we exclude from our testing dataset all methods that cannot be attacked by our method, i.e. all methods without local variables, leaving approximately 40,000 examples. The \pydata{} dataset consists of 50,000 testing examples.

\begin{wrapfigure}{l}{0.5\textwidth}
    \centering
    \begin{subfigure}[t]{.49\linewidth}
    \includegraphics[width=\textwidth]{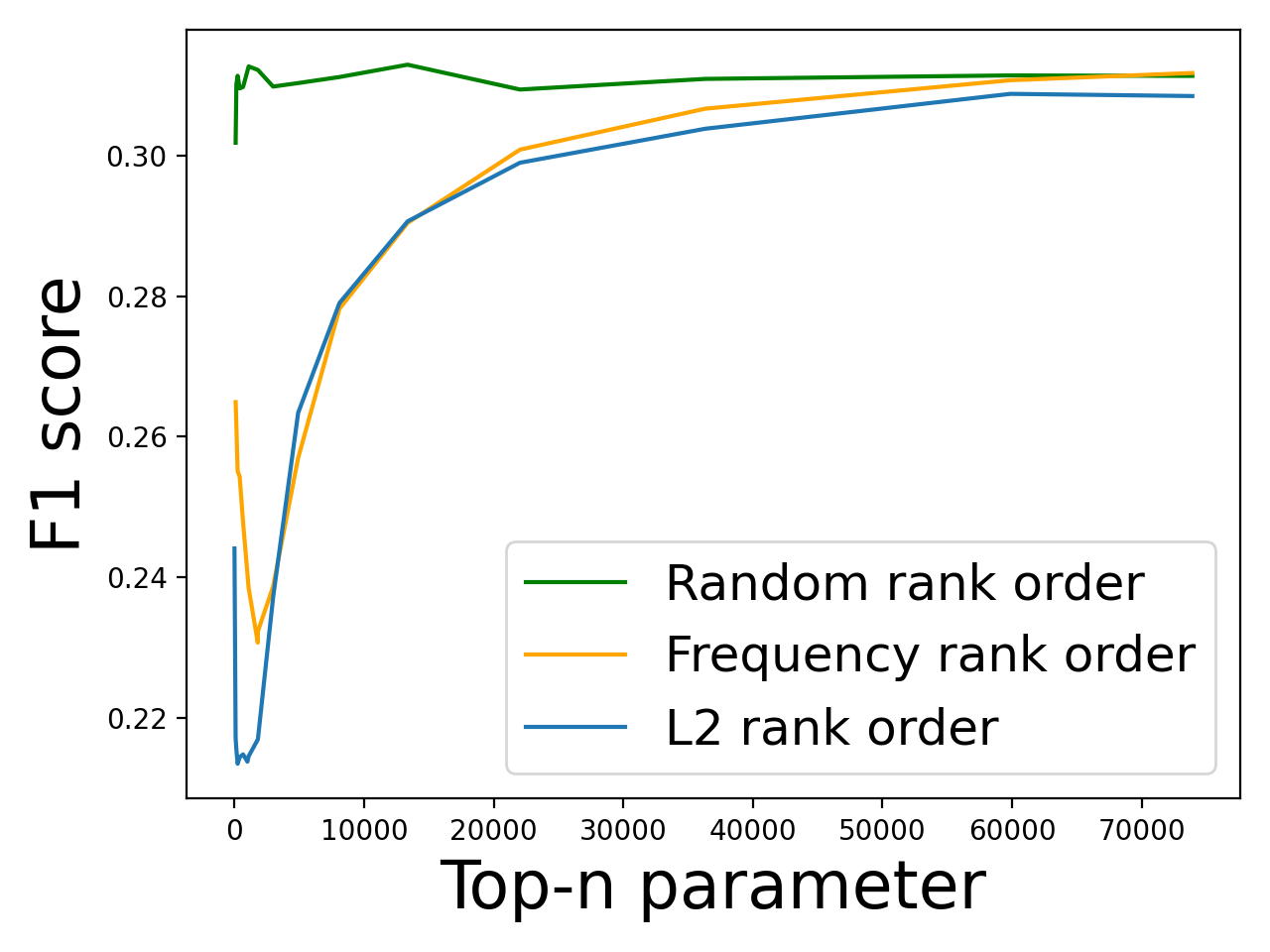}
    \caption{\sname{code2seq/java-small}}
    \end{subfigure}
    \begin{subfigure}[t]{.49\linewidth}
    \includegraphics[width=\textwidth]{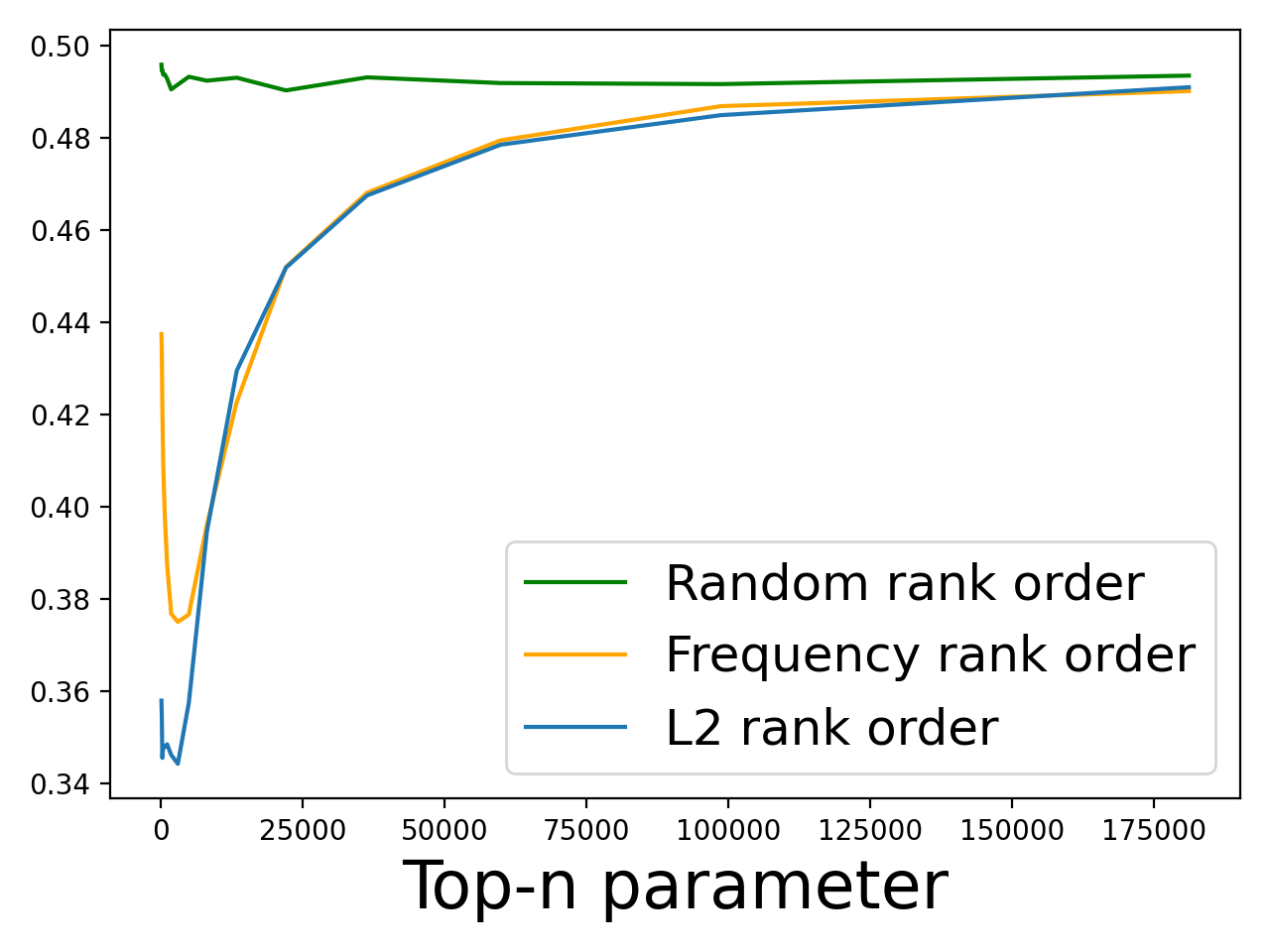}
    \caption{\sname{code2seq/java-med}}
    \end{subfigure}
    \begin{subfigure}[t]{.49\linewidth}
    \includegraphics[width=\textwidth]{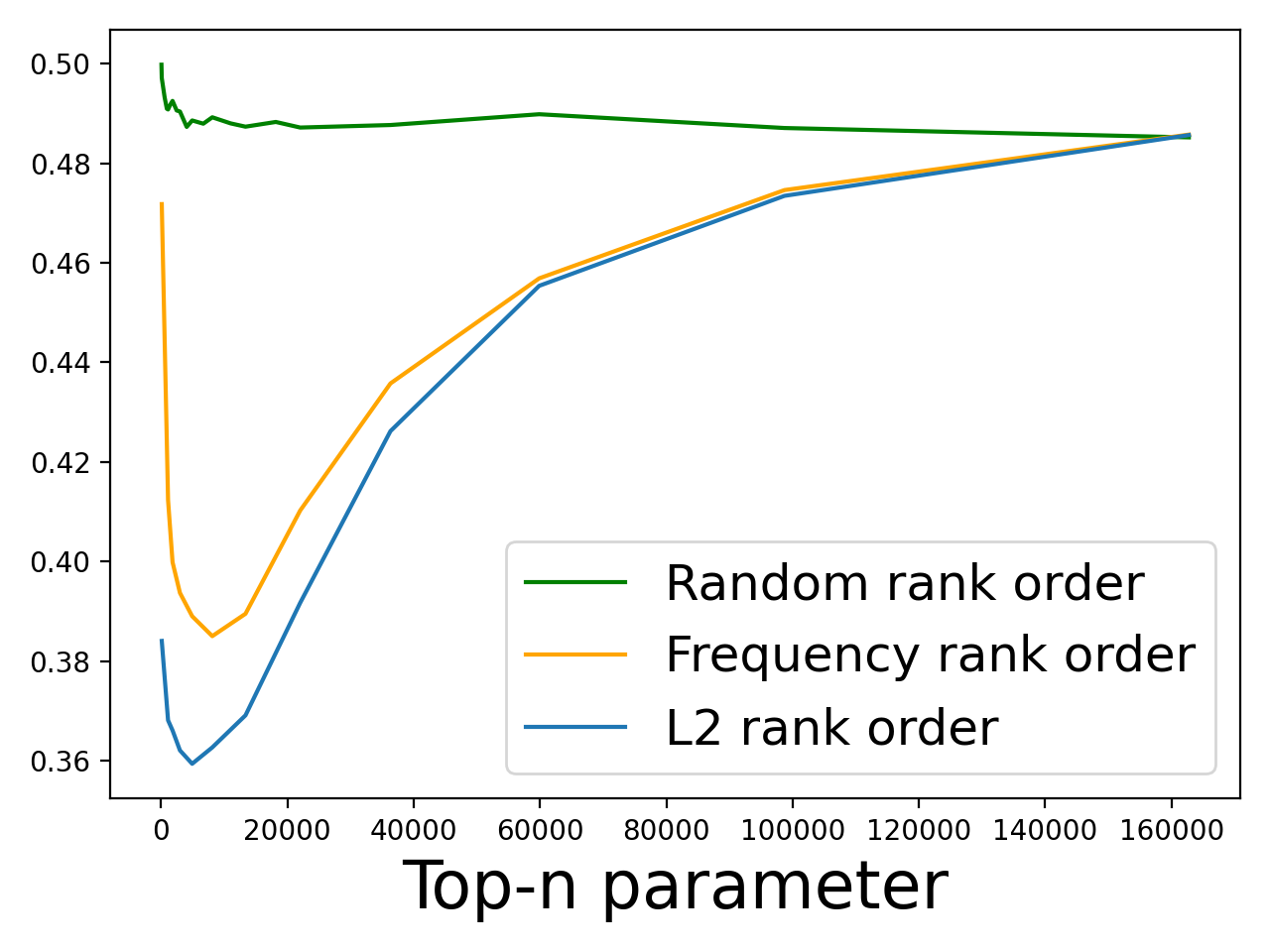}
    \caption{\sname{code2seq/java-large}}
    \end{subfigure}
    \begin{subfigure}[t]{.49\linewidth}
    \includegraphics[width=\textwidth]{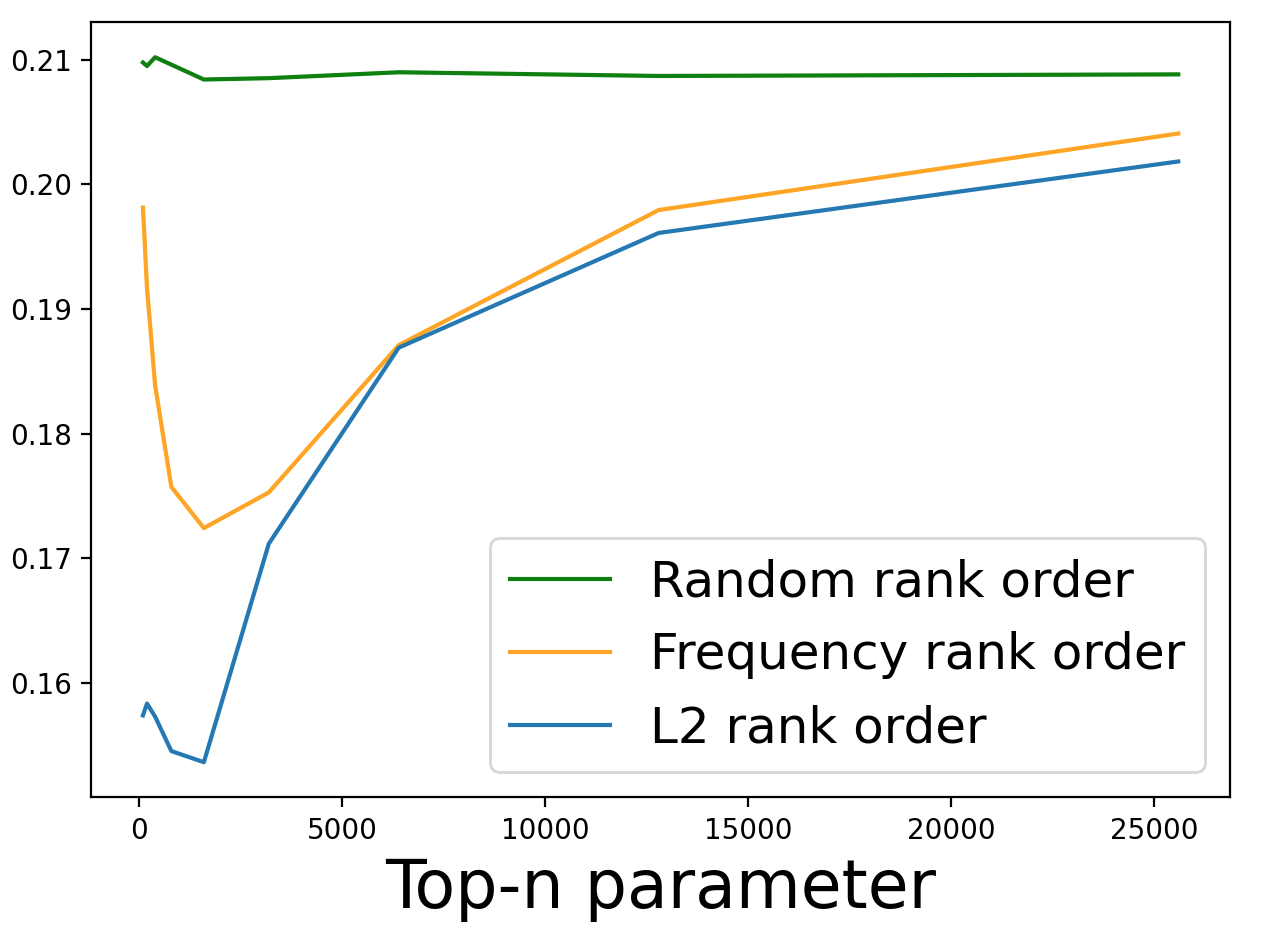}
    \caption{\sname{code2seq/python150k}}
    \end{subfigure}
    \caption{The F1 scores of the \sname{code2seq} models, evaluated on the \javasmall{} or \pydata{} testing dataset that has been adversarially perturbed with \sname{strata} for given top-$n$ parameters, using \textit{5-same} concatenation. A lower F1 score corresponds to a more effective attack.}
    \label{fig:topk_f1}
\end{wrapfigure}

Recall that we choose tokens for the variable replacement attack in three different ways:
\begin{enumerate}[noitemsep,nolistsep]
    \item \textit{All:} Contains all tokens. Note that the number of tokens varies by dataset (Table~\ref{tab:topn});
    \item \textit{Top $n$ by L2 norm} contains tokens for which their L2 norm embedding vectors are the $n$ highest;
    \item \textit{Top $n$ by frequency} contains only the $n$ tokens which occur in the training data with highest frequency.
\end{enumerate}

To obtain optimal thresholds of $n$, we swept the range of possibilities to find the $n$ that minimizes F1 score, i.e., generates the best performing adversarial examples (Figure~\ref{fig:topk_f1}). We present the final values of $n$ in Table~\ref{tab:topn}.

\section{BiLSTM model}

\paragraph{\bilstm{}.} Proposed by \citet{graves2013hybrid} as an extension of the \sname{LSTM} architecture \citep{hochreiter1997long}, the \bilstm{} architecture was originally designed to operate on any sequential data and has been widely adopted in natural language processing. Similarly to natural language, source code can easily be tokenized to act as input for the architecture. Programs are treated as sentences, where each element of the program is considered to be a word. Syntactical constructs are tokenized, and identifiers split into components by camel case and underscores, similarly to \codetoseq{}. By contrast to \codetoseq{}, the \bilstm{} model does not explicitly model the hierarchical and highly nested structure of programs, rather operating on program tokens sequentially. Token embeddings, however, are similarly trained to \codetoseq{} so that they capture semantic structure.

\begin{wrapfigure}{l}{0.33\textwidth}
\includegraphics[width=\linewidth]{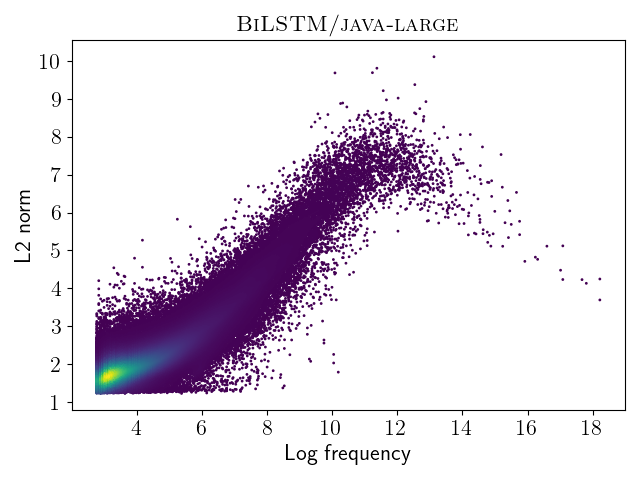}
\caption{The relationship between the frequency of tokens and the L2 norm of the associated token embedding for the \bilstm{} model trained on \javalarge{}.}
\label{fig:bilstm_freq_embed}
\end{wrapfigure}

We find that the \bilstm{} model exhibits the same relationship between the L2 norm of token embedddings and the frequency with which tokens appear in the training dataset (Figure~\ref{fig:bilstm_freq_embed}).

\begin{table*}[ht!]
\centering
\caption{The name of the \codetoseq{} model trained on each dataset, along with optimal values of $n$ for the top-$n$ by frequency and L2 norm vocabularies, and the total number of tokens for each dataset.}
\label{tab:topn}
\begin{tabular}{lllll} \toprule
Dataset    & Model name  & Optimal $n$ by frequency & Optimal $n$ by L2 &  Total tokens \\ \midrule
\javasmall{} & \sname{code2seq/java-small} & $1800$  (2.51\%)             & $1000$ (1.39\%)  & 71766          \\
\javamed{}  & \sname{code2seq/java-med} & $3000$ (1.65\%)              & $3000$ (1.65\%) & 181125           \\
\javalarge{} & \sname{code2seq/java-large} & $10000$ (5.54\%)            & $6000$ (3.33\%) & 180355 \\
\pydata{} & \sname{code2seq/python150k} & $1600$ (1.99\%)            & $1600$ (1.99\%) & 80336 \\
\bottomrule          
\end{tabular}
\end{table*}

\section{Training of token embeddings} 

Both \codetoseq{} and \bilstm{} update token embeddings as frequently as that token appears during training, which is proportional to its representation in the training dataset. However, the training datasets have very large vocabularies consisting not only of standard programming language keywords, but also a huge quantity of neologisms, in the form of variable and method names. The frequency at which different tokens appear in \javalarge{} varies over many orders of magnitude, with the least frequent tokens occurring one time, and the most frequent over $10^8$ times. Source code models need to learn an embedding for each token. The enormity of the token vocabulary in the training set presents a problem for models; with current hardware it would be intractable to store or train this many embeddings. Therefore, both the \codetoseq{} and \bilstm{} models threshold tokens by frequency, replacing the least common tokens with a generic ``unknown'' token. Note that this practice is also common in natural language domains to reduce vocabulary size. Tuning the thresholding parameter may reduce computational cost. We speculate that establishing a correlation between L2 norm and frequency may help identify optimal threshold values to minimize any decrease in accuracy associated with a reduction in the number of token embeddings. For example, we speculate that this parameter should remove as many tokens as possible without removing high-L2-norm tokens.

\begin{wrapfigure}{r}{0.33\textwidth}
    \centering
    \includegraphics[width=\linewidth]{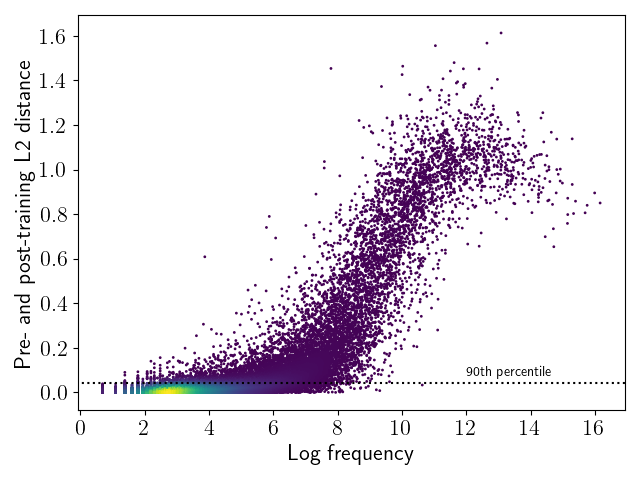}
    \caption{The L2 distance between pre- and post-training embedding vectors for each token, plotted against the log frequency of token in the \javasmall{} training dataset.}
    \label{fig:pre_post_l2}
\end{wrapfigure}

Interested in how this observed correlation relates to the initialization state of the distributed representations, we designed an experiment to test the relationship between pre- and post-training embeddings. Using the \codetoseq{} model, we randomly initialized the embeddings vectors and then measured the L2 distance between each vector's initialization state and its post-training state. This distance is the distance that the embedding vector ``moved" during the training process. We find that high-frequency tokens correspond with embedding vectors that move more significantly during training (Figure~\ref{fig:pre_post_l2}). Since token embedding vectors are updated only as frequently as they are observed by the model during training, this result is largely unsurprising. However, this experiment does imply that the relationship between frequency and L2 norm may have import for model accuracy. Given that 90\% of the randomly initialized embedding vectors are virtually unchanged during training (Figure~\ref{fig:pre_post_l2}), we predict that the presence of these ``low-motility" tokens are not likely to substantially influence the model's prediction, since their values are randomly assigned and thus their position in the embedding space bears very little information to a model that does not over-fit to particular points.

\section{Full comparison with other code2seq attacks}
\label{appendix:comparison}

We have shown that \sname{strata} works well to attack the \sname{code2seq} model and can outperform the attacks by \citet{ramakrishnan_semantic_2020}. Here, we present a comparison by the transformations proposed by \citet{rabin_evaluation_2020}. In order to be able to compare the same metrics, we calculate the percent prediction change, which is the percent of the time that an adversarial change resulted in a change in prediction. A higher percent prediction change indicates a better attack.

In Table~\ref{tab:rabin_comparison}, we compare the performance of our attack to the performance of the transformations generated by \citet{rabin_evaluation_2020} We find that the transformations performed by \citet{rabin_evaluation_2020} result in fewer prediction changes than \sname{strata}. As above, the most effective strategy is our \textit{5-same} attack.

\section{Cross-dataset attacks}
\label{appendix:cross_dataset}

We present two fully black-box attacks that do not require any information about the targeted \codetoseq{} model or dataset:

As a surrogate model, we train a \sname{code2seq} model on any available dataset for the targeted programming language. To obtain adversarial examples from the surrogate, we identify optimal L2 and frequency cutoffs for this model. Using these cutoffs, we construct a vocabulary of the optimal top-$n$ by frequency or by L2 norm. We show that these adversarial examples can be transferred to other models.

\begin{table*}[t]
\centering
\caption{Comparison of the percent of the time the prediction is changed by \sname{strata} and by an attack from \textit{Rabin et al.} Both our attack and the attack from \textit{Rabin et al.} target a \sname{code2seq/java-large}. Note that our model is evaluated on the \javasmall{} testing dataset.}
\label{tab:rabin_comparison}
\begin{tabular}{lc}
\toprule 
Adversarial attack                   & \% prediction change \\ \midrule
Variable renaming (Rabin et al.)     & 47.04                  \\
Boolean Exchange (Rabin et al.)      & 51.43                 \\
Loop Exchange (Rabin et al.)         & 42.51                  \\
Permute Statement (Rabin et al.)     & 43.53                  \\ \midrule
All (\sname{strata}, \textit{5-same}) & 77.3 \\
\textbf{Top 6k by L2 (\sname{strata}, \textit{5-same})}  & \textbf{86.3}  \\
Top 10k by Frequency (\sname{strata}, \textit{5-same}) & 84.5 \\ \bottomrule 
\end{tabular}
\end{table*}

We present the results of the cross-dataset transfer attack proposed in Section~\ref{subsection:dataset_agnostic}. In particular, we generate both frequency and L2 \sname{strata} adversarial examples. We use the L2 norm of the embeddings of \sname{code2seq/java-small}, \sname{code2seq/java-med}, and \sname{code2seq/java-large}, and the token frequencies of \javasmall{}, \javamed{}, and \javalarge{} to construct six different collections of adversarial examples, of which each collection is a perturbation of the \javasmall{} test set. We test each dataset on each model. Tables~\ref{tab:cross_dataset_l2}~and~\ref{tab:cross_dataset_frequency} show the results of the experiments, revealing that while the white-box and known-dataset attacks (the diagonals of the tables) outperform the cross-dataset attacks, the cross-dataset attacks are nonetheless effective. Furthermore, we note that L2-based cross-dataset attacks are more effective than frequency-based cross-dataset attacks, confirming that L2 norms can effectively identify tokens that are high impact in other models. We conclude that \sname{strata} can be performed in a true black-box setting with no information about the model parameters nor the training dataset. The cross-dataset attack is likely effective due to similar distributions of the Java datasets. Similar to word frequencies in natural language corpora, we expect that most Java datasets should have a similar token distributions, and thus \sname{strata} should transfer across models trained on different datasets.

\begin{table*}[ht]
    \centering
    \caption{F1 scores on adversarial data generated by a cross-dataset attack where vocabularies are constructed by using the L2 norm of the embeddings of the \codetoseq{} model trained on the particular dataset. The first column corresponds with the model that is being attacked, and the other columns correspond with the dataset from which the attack is constructed. Lower score means a better attack. All adversarial tokens are concatenated with \textit{5-same}. The boxed scores correspond with the baseline same-dataset attacks.}
    \label{tab:cross_dataset_l2}
    \begin{tabular}{cccc} \toprule
                & \javasmall{} (1k, L2) & \javamed{} (3k, L2) & \javalarge{} (6k, L2) \\ \midrule
    \sname{code2seq/java-small} & \framebox{.214}   & .253             & .290            \\
    \sname{code2seq/java-med} & .359            & \framebox{.349}    & .433            \\
    \sname{code2seq/java-large} & .373            & .361             &  \framebox{.358}  \\ \bottomrule
    \end{tabular}
\end{table*}

\begin{table*}[ht]
\caption{F1 scores on adversarial data generated by a cross-dataset attack where vocabularies are constructed by using the frequency of tokens in the associated training dataset. The first column corresponds with the model that is being attacked, and the other columns correspond with the dataset from which the attack is constructed. Lower score means a better attack. All adversarial tokens are concatenated with \textit{5-same}. The boxed scores correspond with the baseline same-dataset attacks.}
    \label{tab:cross_dataset_frequency}
    \centering
    \begin{tabular}{cccc} \toprule
                & \javasmall{} (1.8k, freq) & \javamed{} (3k, freq) & \javalarge{} (10k, freq) \\ \midrule
    \sname{code2seq/java-small} & \framebox{.235}   & .248             & .279            \\
    \sname{code2seq/java-med} & .381            & \framebox{.377}    & .417            \\
    \sname{code2seq/java-large} & .397            & .392             &  \framebox{.387}  \\ \bottomrule
    \end{tabular}
\end{table*}

\section{Examples of targeted attacks}
\label{appendix:targeted}

\begin{figure*}[ht!]
    \centering
    \begin{subfigure}[tr]{.49\textwidth}
    \includegraphics[width=0.8\textwidth]{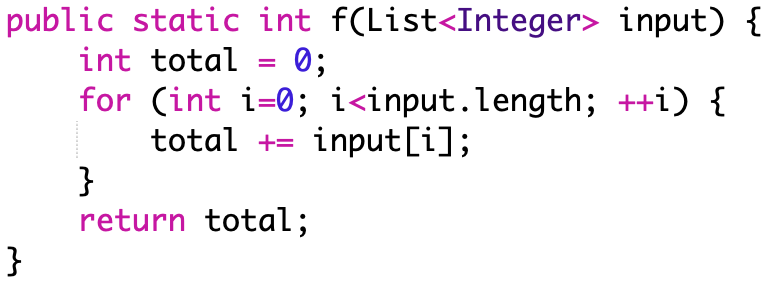}
    \caption{Predicted as \texttt{sum}}
    \end{subfigure}
    \begin{subfigure}[tl]{.49\textwidth}
    \includegraphics[width=0.8\textwidth]{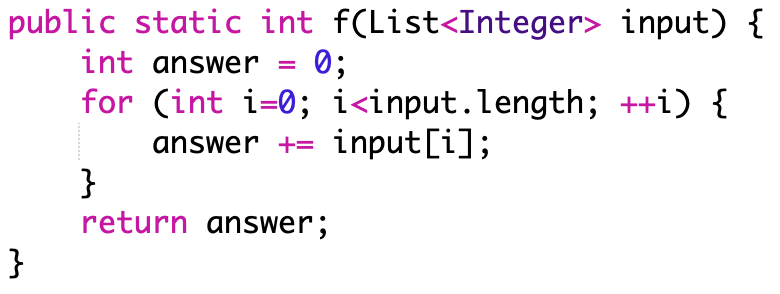}
    \caption{Predicted as \texttt{getInt}}
    \end{subfigure}
    \caption{Changing local variable identifiers affects prediction. \textit{Left}: Original Java method \texttt{sum} computes the sum of a list of integers. \sname{code2seq}, trained on \javalarge{}, predicts that the method name is \texttt{sum}. \textit{Right:} Perturbed Java method with the same behavior, but the local variable \texttt{total} has been replaced with \texttt{answer}, causing \sname{code2seq} to misclassify this method as \texttt{getInt}.}
    \label{fig:change_of_variable_java}
\end{figure*}

To illustrate the effectiveness of \sname{strata} targeted attacks more concretely, we target particular arbitrarily-picked tokens and measure the success rate over the entire testing set (Table~\ref{tab:specific_targeted_attacks}) and find that though effectiveness can vary across different targets, the average effectiveness is quite high.

\begin{table*}[ht!]
\centering
\caption{Effectiveness of targeted attacks on \sname{code2seq/java-large} using uncurated and arbitrarily picked tokens of high embedding vector L2 norm, sorted by \% success.}
\label{tab:specific_targeted_attacks}
\begin{tabular}{lc|lc|lc|lc} \toprule
Target   & \% & Target & \% & Target & \% & Target & \% \\ \midrule
tournament & 86.3       & ftp      & 53.4       & concat   & 41.5       & combat   & 26.6       \\
redaction  & 70.5       & podam    & 45.3       & eternal  & 37.3       & thursday & 23.9       \\
outliers   & 62.4       & wind     & 44.7       & orderby  & 32.8       & girth    & 21.5       \\
mission    & 56.8       & weld     & 42.4       & reentry  & 29.8       & ixor     & 16.4      \\ \bottomrule
\end{tabular}
\end{table*}

\section{Transferability}

In this section we show that gradient-based adversarial training as proposed by \citet{ramakrishnan_semantic_2020} is not effective at improving robustness to \sname{strata}.  We test a \sname{code2seq} model that has been trained to be robust to the gradient-based adversarial examples proposed by \citet{ramakrishnan_semantic_2020}, and find that the model is indeed robust to the gradient-based token-replacement perturbations. However, neither the original nor the robust model are impervious to perturbations produced by \sname{strata} (Table~\ref{tab:transferability}). This result confirms that \sname{strata} can effectively target models that are robust to some gradient-based perturbations; therefore it is a useful tool when hardening models of code, even when gradient-based perturbations are also being used.

\begin{table*}[ht]
\centering
\caption{F1 scores of a non-robust and robust \sname{code2seq/java-small} model on gradient-based adversarial perturbations and \sname{strata} perturbations.}
\label{tab:transferability}
\begin{tabular}{llll} 
\toprule
Model                & No perturbations (F1) & Gradient perturbations (F1) & \sname{strata} perturbations (F1) \\ \midrule
Original &         .363              &                 .243                      &          \textbf{.212}                 \\
Robust   &            .367           &                  .342                    &          \textbf{.240}                 \\ \bottomrule
\end{tabular}

\end{table*}

\section{Computationally inexpensive}
\label{appendix:efficient}

Current alternative methods for attacking models of source code with comparable results involve either an extensive search for optimal transformations or gradient-based optimization for token replacement, or a combination of the two \citep{yefet_adversarial_2020, ramakrishnan_semantic_2020, rabin_evaluation_2020}. Extensive searches are inherently computationally expensive and, in the case of gradient-based optimization, oftentimes require a GPU for efficient implementation. \sname{strata}, however, can be implemented to run quickly on even CPU-only machines. After an initial pre-processing step to mark local variable names for easy replacement which took less than five minutes on our 24-core CPU-only machine, we were able to construct adversarial examples using \sname{strata} on a dataset of 20,000 within seconds. The analogous gradient-based method proposed by \citet{ramakrishnan_semantic_2020} took multiple hours on the same machine. The combined speed and effectiveness of \sname{strata} will allow researchers to quickly harden their models against adversarial attacks with efficient large-scale adversarial training.

\section{Limitations}

Our work does have some limitations. While we find that choosing tokens that have high-L2-norm embeddings to construct identifier replacements substantially improves the impact of variable replacement attacks on model accuracy when compared with random token selection on the \codetoseq{} model, we do not observe any increase in attack performance on the \bilstm{} model. We hypothesize that the fact that \codetoseq{} splits identifiers into separate tokens and then sums the constituent token embeddings to construct the corresponding identifier embedding may cause higher-impact tokens to have an even larger embedding L2 norm. So, while the \codetoseq{} vulnerability relates to the correlation between frequency and embedding L2 norm, we have not yet uncovered the whole story.

\end{document}